# Are GNNs Worth the Effort for IoT Botnet Detection? A Comparative Study of VAE-GNN vs. ViT-MLP and VAE-MLP Approaches

Hassan Wasswa[1], Hussein Abbass[1], Timothy Lynar[1]
[1]School of Systems and Computing, University of New South Wales, Canberra, ACT, Australia
h.wasswa@unsw.edu.au, h.abbass@unsw.edu.au, t.lynar@unsw.edu.au,

*Abstract*—Due to the exponential rise in IoT-based botnet attacks, researchers have explored various advanced techniques for both dimensionality reduction and attack detection to enhance IoT security. Among these, Variational Autoencoders (VAE), Vision Transformers (ViT), and Graph Neural Networks (GNN), including Graph Convolutional Networks (GCN) and Graph Attention Networks (GAT), have garnered significant research attention in the domain of attack detection. This study evaluates the effectiveness of four state-of-the-art deep learning architectures for IoT botnet detection: a VAE encoder with a Multi-Layer Perceptron (MLP), a VAE encoder with a GCN, a VAE encoder with a GAT, and a ViT encoder with an MLP. The evaluation is conducted on a widely studied IoT benchmark dataset—the N-BaIoT dataset for both binary and multiclass tasks. For the binary classification task, all models achieved over 99.93% in accuracy, recall, precision, and F1- score, with no notable differences in performance. In contrast, for the multiclass classification task, GNN-based models showed significantly lower performance compared to VAE-MLP and ViT- MLP, with accuracies of 86.42%, 89.46%, 99.72%, and 98.38% for VAE-GCN, VAE-GAT, VAE-MLP, and ViT-MLP, respectively.

## I. INTRODUCTION

The rapid proliferation of Internet of Things (IoT) technology has resulted in an exponential increase in the number of IoT devices. At present, IoT is widely adopted by both individual users and professionals across various domains, ranging from small businesses to large enterprises. However, a significant proportion of these devices remain weakly secured, often relying on default login credentials provided by manufacturers. Moreover, the inherent resource constraints of IoT networks—such as limited memory, processing power, and bandwidth—pose significant challenges to the implementation of conventional security mechanisms, including antivirus software, encryption systems, and intrusion detection systems. As a result, IoT networks present an attractive attack surface for cybercriminals who exploit large volumes of weakly secured IoT devices to construct IoT botnets.

By aggregating the computational resources of hundreds of thousands of compromised IoT devices, attackers can launch large-scale distributed attacks such as Distributed Denial of Service (DDoS), distributed click fraud, and the dissemination of widespread spam campaigns. In response to this growing threat, and motivated by the recent successes of deep learning algorithms in domains such as computer vision [1], [2], pattern recognition [3]–[5], and natural language processing (NLP) [6]–[8], significant research efforts have been directed toward developing robust and efficient IoT botnet detection models to enhance network security.

A notable advancement in this domain is the increasing adoption of Variational Autoencoder (VAE)-based and attention mechanism-based methods. These approaches commonly involve training detection models on latent space representations derived from high-dimensional feature inputs processed through VAE or Vision Transformer (ViT) encoders. However, as noted in studies such as [9], [10], a major limitation of these models lies in their treatment of each flow instance as an independent entity, thereby neglecting the inherent interdependencies among instances. To mitigate this issue, recent works have proposed incorporating Graph Neural Network (GNN) techniques, which are capable of modeling relationships among NetFlow instances, thereby improving the effectiveness of attack detection.

This study aimed to investigate a critical research question: *Are GNNs worth the effort for IoT botnet detection?* To this end, the classification performance of four advanced deep learning frameworks—VAE-GCN, VAE-GAT, VAE-MLP, and ViT-MLP—was systematically evaluated using the N-BaIoT dataset. The methodology involved projecting the high-dimensional feature space into a lower-dimensional latent representation using encoder components of: a VAE for the VAE-GCN, VAE-GAT, and VAE-MLP models, and a ViT encoder for the ViT-MLP model. Subsequently, the classification algorithms—GCN, GAT, and MLP—were trained on the resulting latent representations for both binary (benign vs. attack) and multi-class traffic classification tasks. Model performance was assessed using standard evaluation metrics, namely accuracy, precision, recall, and F1-score. The results indicated that, for binary classification, none of the models demonstrated a clear advantage. However, in the multi-class classification scenario, GNN-based models (VAE-GCN and VAE-GAT) exhibited significantly lower performance compared to both the VAE-MLP and ViT-MLP models.

The rest of this paper is structured as follows. Section II reviews existing literature on the integration of graph neural networks with attention mechanisms for anomaly detection. Section III outlines the proposed framework, provides an overview of the dataset, and describes the techniques employed. Section IV presents and interprets the experimental results, and Section V concludes the study.

## II. RELATED WORK

This section presents a review of prior work in the context of GNNs and Attention mechanisms for attack detection.

### A. Prior studies deploying attention mechanism for attack detection

Studies [11], [12] introduced a method aimed at adapting transformer models—specifically the ViT—for the detection of IoT botnet attacks using network flow packet data. The method involved extracting features from ".pcap" files and transforming each instance into a 1-channel 2D image format to facilitate ViT-based classification. Additionally, the ViT architecture was modified to support the integration of various classifiers beyond the MLP originally used in the ViT framework. Classifiers such as the conventional feedforward Deep Neural Network (DNN), LSTM, and Bidirectional LSTM were evaluated, and they demonstrated competitive performance in terms of precision, recall, and F1-score for multiclass attack detection across two IoT attack datasets.

Study [13] investigated how varying latent dimensions affect the performance of deep learning classifiers trained on low-dimensional representations of IoT botnet traffic data. By comparing encoder components from ViT and VAE, the study aimed to determine which model more effectively projected high-dimensional structured datasets into latent space. Using the N-BaIoT and CICIoT2022 datasets, the results showed that classifiers trained on VAE-based embeddings outperformed those based on ViT across all performance metrics, highlighting the VAE's superior suitability for structured, NetFlow data.

The study in [14] aimed to detect cross-site scripting (XSS) attacks by comparing a Transformer-based model with RNN architectures (LSTM and GRU), using tokenization and NLP techniques such as word2vec (skip-gram), TF-IDF, and CBOW to generate word feature vectors from segmented URL components. In [15], the authors focused on classifying IoT botnet malware using a Transformer model by disassembling malware opcodes (Mirai, Gafgyt, and unknown samples), extracting instruction sequences, applying CBOW for word embeddings, and using a feedforward neural network for final classification. The study in [16] deployed the ViT for detecting image spoofing attacks in face recognition systems, using patch-wise data augmentation techniques—including intra-class patch mixing, live patch masking, and patch cutout—on datasets such as SiW [17] and Oulu-NPU [18]. Finally, [19] proposed a DGA domain detection scheme using the CANINE Transformer model, where domain names were tokenized into character sequences and converted into input representations (input IDs, attention masks, and token type IDs) to fine-tune the model for distinguishing between benign and DGA-generated domain names.

### B. Prior studies deploying GNN-based models for attack detection

Recent research has explored hybrid approaches that integrate GNNs with transformer-based architectures to improve anomaly detection by leveraging the strengths of both models. Study [20] proposed a model combining a GNN for capturing complex structural relationships with a transformer to process long-range dependencies for intrusion detection. Study [10] developed a framework integrating a graph convolutional network and attention mechanisms to identify coordinated behaviors in IoT device graphs for botnet detection. Study [21] focused on attack detection in EV charging stations by constructing graphs from hardware logs and using a GNN alongside a transformer to model feature correlations and interactions. Study [22] introduced AJSAGE, which enhances the GraphSAGE model with an attention mechanism to detect abnormal traffic nodes in network attack graphs. These methodologies underscore the potential of GNN-based models across various cybersecurity applications.

## III. METHODOLOGY

The GNN architectures evaluated in this study are the GCN and the GAT. The GAT model incorporates an attention mechanism into the GCN framework, enabling it to capture long-range feature dependencies and inter-instance relationships more effectively. However, training a GNN on NetFlow instances necessitates converting the structured traffic data into a graph-structured format. To facilitate this, each instance in the dataset was treated as a node, and neighborhood was established using the kNN algorithms, (with *n_neighbors = 3* and *metric = "euclidean"*), to construct the graph data structure. Moreover, constructing a graph directly from the high-dimensional NetFlow data can lead to a complex graph structure, demanding substantial system memory and computational resources. To mitigate this, the high-dimensional NetFlow instances were projected into an 8-dimensional latent space using a VAE encoder, and the graph was constructed on this low-dimensional representation. Fig. 1-(a) illustrates the GNN-based IoT botnet detection framework employed in this study.

To compare the GNN models—GCN and GAT—with other state-of-the-art approaches, two additional models, VAE-MLP and ViT-MLP, were trained on the same dataset. Similar to the GNN models, the high-dimensional input data is projected into an 8-dimensional latent space before classification. In the VAE-MLP model, a VAE is trained with a latent space of dimension 8, while in the ViT-MLP model, the ViT encoder is configured to output an 8-dimensional representation. A MLP is then stacked on top of the ViT encoder to perform traffic classification. The overall methodology for VAE/ViT- based IoT botnet detection adopted in this study is illustrated in Fig. 1-(b).

### A. Variational Autoencoder

The Variational Autoencoder [23] was introduced to address the lack of regularization in the latent space often encountered in traditional Autoencoders (AEs). By leveraging Bayesian variational inference, VAEs learn parameters for both the encoder and decoder to approximate the underlying data distribution. This is achieved by introducing a latent variable and optimizing a surrogate network to estimate the posterior distribution. The training objective focuses on maximizing the Evidence Lower Bound (ELBO), which indirectly minimizes the Kullback–Leibler divergence between the approximate and true posterior distributions, thereby facilitating more structured and meaningful latent representations.

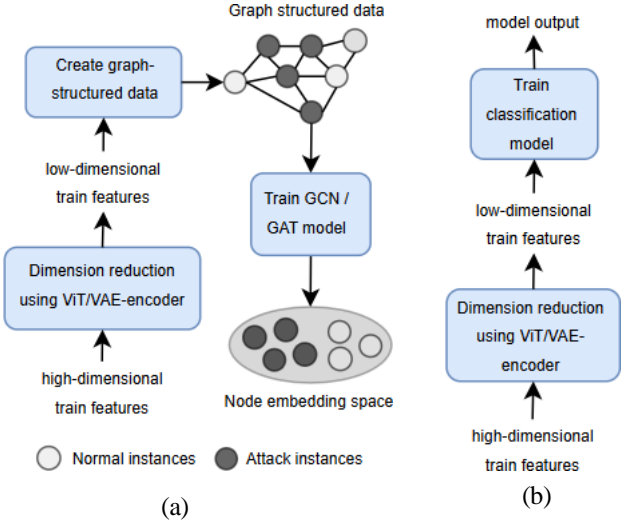

Fig. 1: Design flow of the two state-of-the-art model architectures: (a) GNN-based (b) VAE/ViT-based model

### B. Vision Transformer

The Vision Transformer, introduced in [24], extends the self-attention mechanism of the original Transformer architecture—initially developed for natural language processing (NLP)—to address computer vision challenges. In contrast to conventional convolutional neural networks (CNNs), ViT divides an image into uniform, fixed-size patches and applies self-attention to model the inter-patch relationships. This design allows the model to effectively capture global context and long-range interactions across the image. Each patch is flattened into a one-dimensional vector and passed through a linear projection to form the input embeddings. To preserve spatial arrangement, positional embeddings are added to the patch vectors, enabling the model to interpret their relative locations within the image.

### C. Dataset

The performance evaluation of the models was done using the N-BaIoT dataset [25]. The testbed comprised nine commercial IoT devices each subjected to two prominent IoT botnet malware families: Mirai and BashLite. A total of 2,482,470 instances were retained after the removal of duplicate row entries. The instances were distributed across ten traffic classes as: {"Normal": 513,497 (21.52%), "mirai_udp": 555,973 (23.30%), "mirai_syn": 317,115 (13.29%), "mirai_ack": 280,144 (11.74%), "mirai_scan": 256,151 (10.74%), "gafgyt_udp": 107,665 (4.51%), "gafgyt_combo": 62,213 (2.61%), "gafgyt_junk": 31,293 (1.31%), "gafgyt_scan": 31,087 (1.30%), "mirai_udpplain": 230,508 (9.66%)}.

To enable compatibility between the ViT encoder model and the NetFlow dataset, each data instance is reformatted to resemble a single-channel 2D image. Specifically, an instance $x \in R^d$ is reshaped into $x' \in R^{r \times k \times 1}$ such that $r \times k = d$. The transformed instance $x'$ is then partitioned into equally sized image patches prior to being input into the ViT architecture. Given that the N-BaIoT dataset comprises 115 independent features, each instance is reshaped into a $5 \times 23 \times 1$ image, from which 23 uniform patches of size $5 \times 1 \times 1$ are extracted for processing by the ViT model. Fig. 2 illustrates the ViT-MLP architecture deployed in this study.

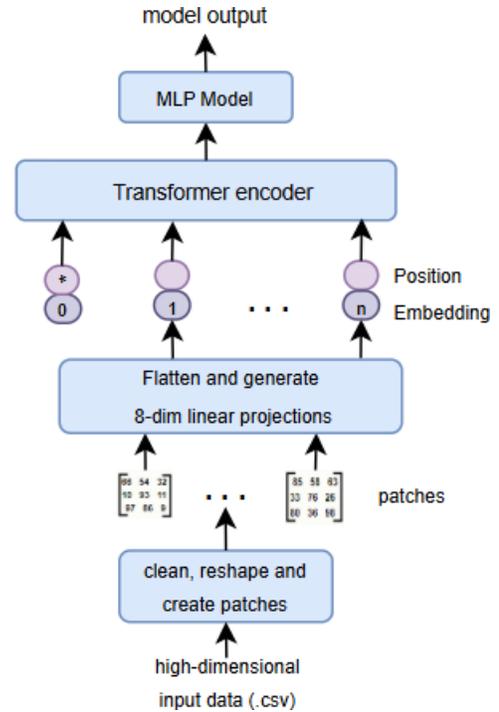

Fig. 2: ViT-MLP architecture

### D. Graph Neural Networks

Traditional deep learning models such as CNNs and RNNs are effective for structured data like sequences and images but are limited when applied to graph-structured data. Graph Neural Networks, first introduced in [26], overcome this limitation by enabling the modeling of entities and their relationships in graph form. This capability is particularly valuable for domains involving complex and dynamic structures, such as molecular biology, materials science, and NetFlow analysis.

In the context of NetFlow analysis, GNNs represent traffic instances as nodes and define edges based on relationships such as Euclidean distance, cosine similarity or IP address connections. Through iterative message passing, GNNs learn node embeddings that capture both individual features and local structural information, thereby enabling accurate classification of network traffic into corresponding attack types.

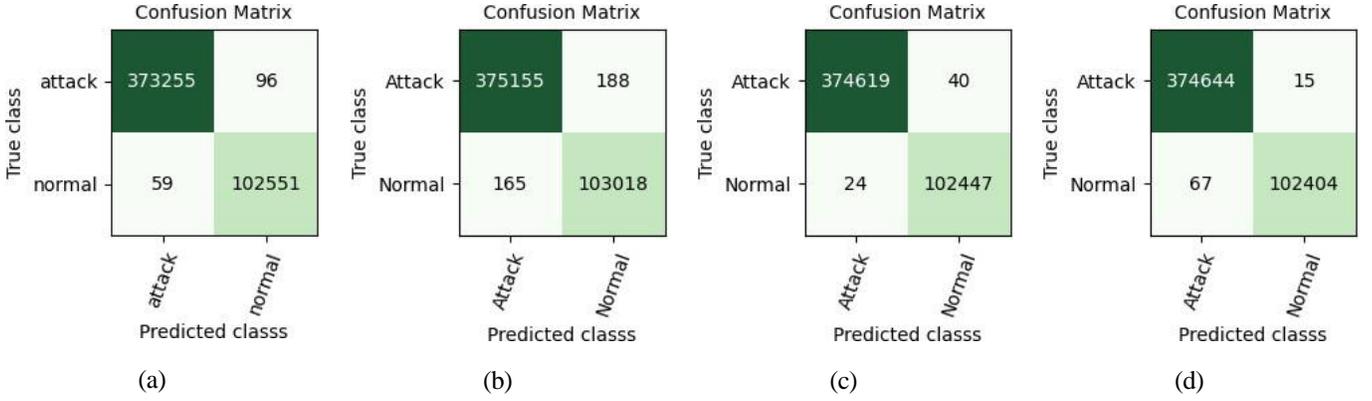

Fig. 3: Confusion matrix: (a) VAE-GCN (b) VAE-GAT (c) VAE-MLP and (d) ViT-MLP model

### E. Model Training and evaluation

The dataset was randomly divided into training and testing sets with an 80%-20% split for model training and evaluation. Additionally, 10% of each training batch was randomly reserved for validation during every training epoch. The ReLU activation function was utilized for neuron activation, and the Adam optimizer was applied with a learning rate of 0.001. Across all configurations, each model was trained for 20 epochs using a batch size of 128.

## IV. RESULTS AND DISCUSSION

### A. Binary classification results

The performance of the models in distinguishing attack traffic from benign traffic was evaluated and compared. All models achieved exceptionally high results, with classification accuracy, recall, precision, and F1-score each exceeding 99.93%, and no model demonstrated a significant advantage over the others, as shown in Fig. 4. Additionally, Fig. 3 provides a visual representation of the distribution of test class instances across the two traffic categories for each of the four models.

### B. Multiclass classification results

This section reports the findings for the multiclass task using accuracy, weighted averages of precision, recall, and F1-score, as well as confusion matrices for the ten-class classification problem. The results demonstrate that the VAE-GCN model achieved an accuracy of 86.42%, VAE-GAT reached 89.46%, VAE-MLP attained 99.72%, and ViT-MLP obtained 98.38%. A visual comparison of the models' performance across accuracy, precision, recall, and F1-score is provided in Fig. 5.

The results indicate that the GNN-based models—GCN and GAT—perform significantly worse than the other advanced models—VAE-MLP and ViT-MLP. This performance gap is further evidenced by the distribution of test instances across the ten traffic classes, as illustrated by the confusion matrices in Fig. 6 where the VAE-GCN and VAE-GAT models exhibit a higher rate of misclassification compared to the VAE-MLP and ViT-MLP models.

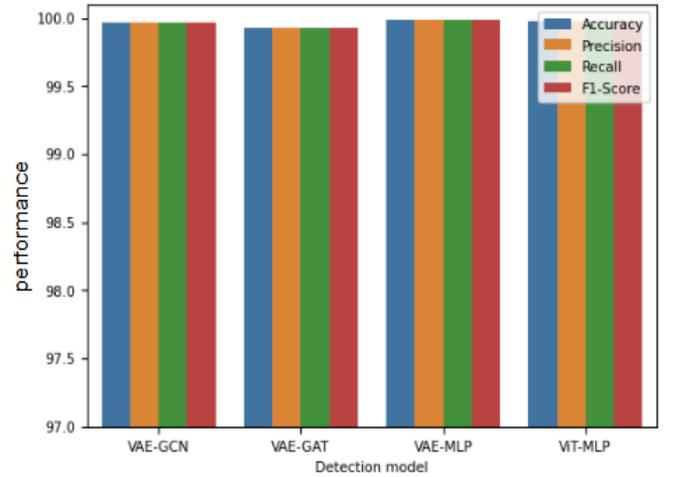

Fig. 4: Performance comparison for binary classification

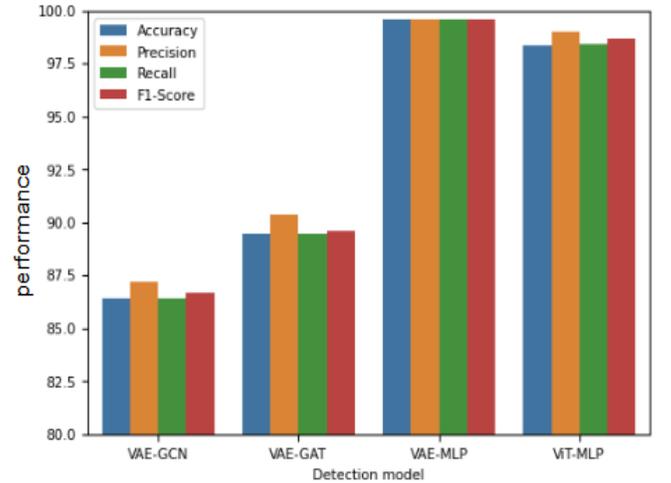

Fig. 5: Performance comparison for multiclass classification

### C. Computational cost

**VAE Time Complexity**: VAEs use dense layers in both the encoder and decoder, each with time complexity $O(d_{in} \cdot d_{out})$. With $a$ encoder and $b$ decoder layers, and constant-time latent sampling $O(1)$, the overall time complexity is $O((a+b) \cdot d_{in} \cdot$

5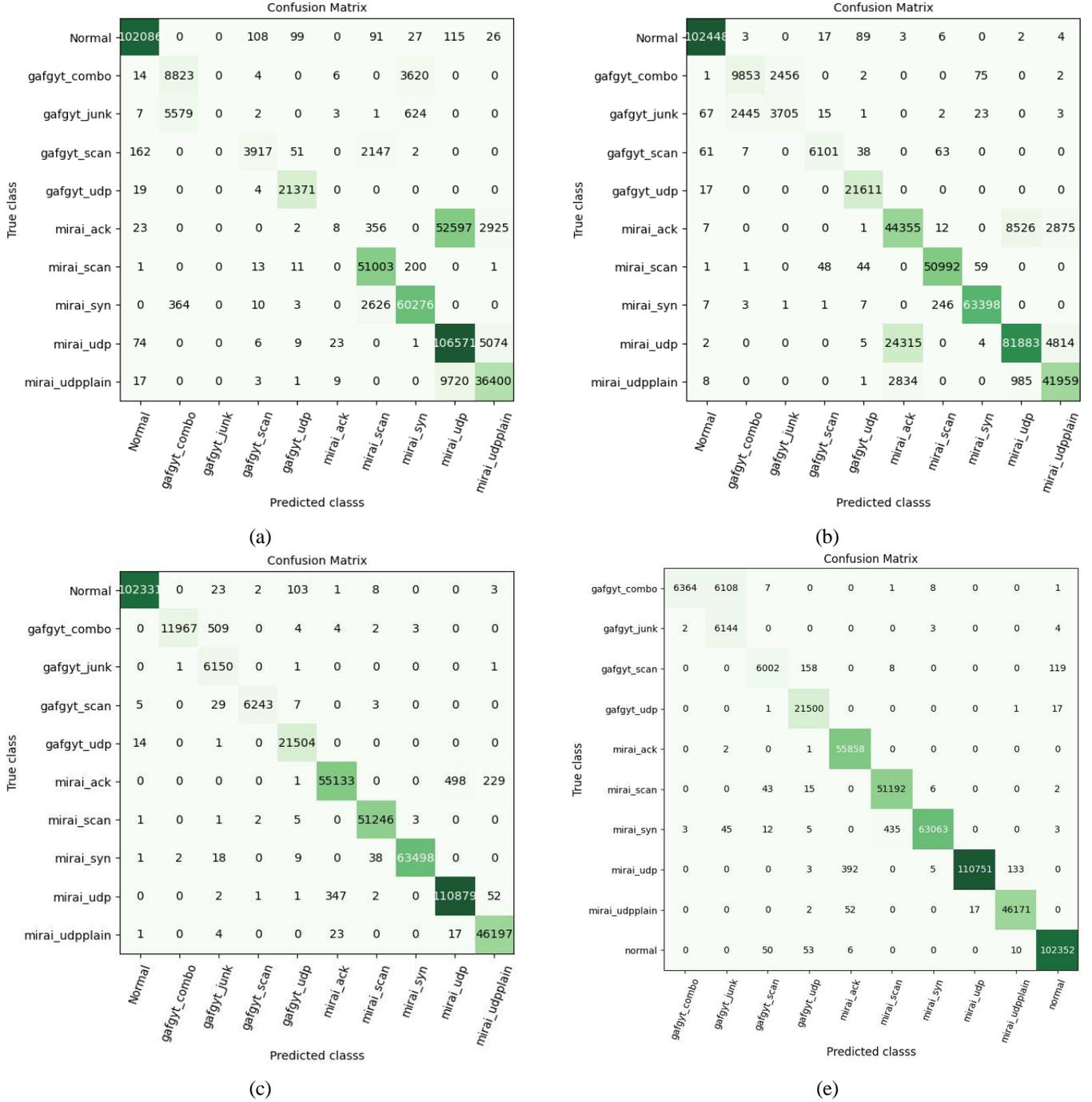

Fig. 6: Confusion matrix: (a) VAE-GCN model (b) VAE-GAT model (C) VAE-MLP model (d) ViT-MLP model

$d_{out} + 1)$, or $O(c \cdot d_{in} \cdot d_{out} + 1)$ where $c = a + b$.

**ViT Time Complexity**: The self-attention mechanism in each Transformer encoder layer dominates the ViT's time complexity. For $p$ patches of dimension $d$, self-attention costs $O(p^2 \cdot d)$ and the feedforward network adds $O(p \cdot d^2)$, resulting in $O(p^2 \cdot d + p \cdot d^2)$ per layer. For $n$ layers, the total complexity is $O(n \cdot (p^2 \cdot d + p \cdot d^2))$.

**Graph data construction time complexity**: The time complexity of constructing a sparse graph with $N$ nodes (each of dimension $D$) and $E$ edges is $O(ND^2 + ED)$.

**GCN Time Complexity**: The time complexity of a Graph Convolutional Network (GCN) layer depends on the number of nodes $N$, the number of edges $E$, and the input and out feature dimensions, $d_{in}$, and $d_{out}$ respectively. For a GCN with $n$ layers, the overall time complexity is given as $O(n(E \cdot d_{in} + N \cdot d_{in} \cdot d_{out}))$.

**GAT model time complexity**: The time complexity of the GAT model with $n$ GATConv layers, $N$ nodes, $E$ edges, input dimension $D$, attention heads $H$, and output feature size $K$ per head is $O(n(NDK + HEK))$.

Table I summarizes the computational complexity associated with using each of the detection frameworks. The analysis clearly shows that GNN-based models are more computationally intensive, while VAE-MLP demonstrates the lowest computational cost.

TABLE I: Computational cost analysis

| Framework | Time Complexity |
|---|---|
| VAE-GCN | $O(c(\cdot d_{in} \cdot d_{out})) + O(ND^2 + ED) + O(n(E \cdot d_{in} + N \cdot d_{in} \cdot d_{out}))$ |
| VAE-GAT | $O(c(\cdot d_{in} \cdot d_{out})) + O(ND^2 + ED) + O(n(NDK + HEK))$ |
| VAE-MLP | $O(c(\cdot d_{in} \cdot d_{out}) + 1) + O(n(\cdot d_{in} \cdot d_{out}))$ |
| ViT-MLP | $O(n \cdot (p^2 \cdot d + p \cdot d^2)) + O(n(\cdot d_{in} \cdot d_{out}))$ |

## V. Conclusion

This study evaluated the performance of four deep learning architectures—VAE-GCN, VAE-GAT, VAE-MLP, and ViT-MLP—for IoT botnet detection using the N-BaIoT benchmark dataset. Results show that while all models performed exceptionally well on the binary classification task, GNN-based models lagged behind VAE-MLP and ViT-MLP in the multiclass setting. Additionally, computational analysis revealed that GNN-based models are significantly more resource- intensive than VAE-MLP. The combination of lower classification performance and higher computational cost makes GNN-based models less suitable for deployment in resource- constrained IoT environments. Future work should focus on enhancing the representational power and generalization of GNNs to better address the complexity and evolving nature of IoT threats.

## References

[1] M. Yuzhan, H. Abdullah Jalab, W. Hassan, D. Fan, and M. Minjin, "Recaptured image forensics based on image illumination and texture features," in *Proceedings of the 2020 4th International Conference on Video and Image Processing*, 2020, pp. 93–97.

[2] H. Wasswa and A. Serwadda, "The proof is in the glare: On the privacy risk posed by eyeglasses in video calls," in *Proceedings of the 2022 ACM on International Workshop on Security and Privacy Analytics*, ser. IWSPA '22. New York, NY, USA: Association for Computing Machinery, 2022, p. 46–54.

[3] S. Ali, R. Ghazal, N. Qadeer, O. Saidani, F. Alhayan, A. Masood, R. Saleem, M. A. Khan, and D. Gupta, "A novel approach of botnet detection using hybrid deep learning for enhancing security in iot networks," *Alexandria Engineering Journal*, vol. 103, pp. 88–97, 2024.

[4] H. Wasswa, T. Lynar, and H. Abbass, "Enhancing iot-botnet detection using variational auto-encoder and cost-sensitive learning: A deep learning approach for imbalanced datasets," in *2023 IEEE Region 10 Symposium (TENSYMP)*. IEEE, 2023, pp. 1–6.

[5] G. Casqueiro, S. E. Arefin, T. Ashrafi Heya, A. Serwadda, and H. Wasswa, "Weaponizing iot sensors: When table choice poses a security vulnerability," in *2022 IEEE 4th International Conference on Trust, Privacy and Security in Intelligent Systems, and Applications (TPS-ISA)*, 2022, pp. 160–167.

[6] R. Amin, R. Gantassi, N. Ahmed, A. H. Alshehri, F. S. Alsubaei, and J. Frnda, "A hybrid approach for adversarial attack detection based on sentiment analysis model using machine learning," *Engineering Science and Technology, an International Journal*, vol. 58, p. 101829, 2024.

[7] A. Nanyonga, H. Wasswa, U. Turhan, K. Joiner, and G. Wild, "Exploring aviation incident narratives using topic modeling and clustering techniques," in *2024 IEEE Region 10 Symposium (TENSYMP)*, 2024, pp. 1–6.

[8] W. S. Ismail, "Threat detection and response using ai and nlp in cybersecurity," *J. Internet Serv. Inf. Secur.*, vol. 14, no. 1, pp. 195–205, 2024.

[9] L. Lin, Q. Zhong, J. Qiu, and Z. Liang, "E-gracl: an iot intrusion detection system based on graph neural networks," *The Journal of Supercomputing*, vol. 81, no. 1, p. 42, 2025.

[10] J. Kumar *et al.*, "Grma-cnn: Integrating spatial-spectral layers with modified attention for botnet detection using graph convolution for securing networks." *International Journal of Intelligent Engineering & Systems*, vol. 18, no. 1, 2025.

[11] H. Wasswa, T. Lynar, A. Nanyonga, and H. Abbass, "Iot botnet detection: Application of vision transformer to classification of network flow traffic," in *2023 Global Conference on Information Technologies and Communications (GCITC)*, 2023, pp. 1–6.

[12] H. Wasswa, H. A. Abbass, and T. Lynar, "Resdnvit: A hybrid architecture for netflow-based attack detection using a residual dense network and vision transformer," *Expert Systems with Applications*, vol. 282, p. 127504, 2025.

[13] H. Wasswa, A. Nanyonga, and T. Lynar, "Impact of latent space dimension on iot botnet detection performance: Vae-encoder versus vit-encoder," in *2024 3rd International Conference for Innovation in Technology (INOCON)*, 2024, pp. 1–6.

[14] B. Peng, X. Xiao, and J. Wang, "Cross-site scripting attack detection method based on transformer," in *2022 IEEE 8th International Conference on Computer and Communications (ICCC)*. IEEE, 2022, pp. 1651–1655.

[15] X. Hu, R. Sun, K. Xu, Y. Zhang, and P. Chang, "Exploit internal structural information for iot malware detection based on hierarchical transformer model," in *2020 IEEE 19th International Conference on Trust, Security and Privacy in Computing and Communications (TrustCom)*, 2020, pp. 927–934.

[16] K. Watanabe, K. Ito, and T. Aoki, "Spoofing attack detection in face recognition system using vision transformer with patch-wise data augmentation," in *2022 Asia-Pacific Signal and Information Processing Association Annual Summit and Conference (APSIPA ASC)*. IEEE, 2022, pp. 1561–1565.

[17] Y. Liu, A. Jourabloo, and X. Liu, "Learning deep models for face anti-spoofing: Binary or auxiliary supervision," in *Proceedings of the IEEE conference on computer vision and pattern recognition*, 2018, pp. 389–398.

[18] Z. Boulkenafet, J. Komulainen, L. Li, X. Feng, and A. Hadid, "Oulu-npu: A mobile face presentation attack database with real-world variations," in *2017 12th IEEE international conference on automatic face & gesture recognition (FG 2017)*. IEEE, 2017, pp. 612–618.

[19] B. Gogoi and T. Ahmed, "Dga domain detection using pretrained character based transformer models," in *2023 IEEE Guwahati Subsection Conference (GCON)*. IEEE, 2023, pp. 01–06.

[20] H. Zhang and T. Cao, "A hybrid approach to network intrusion detection based on graph neural networks and transformer architectures," in *2024 14th International Conference on Information Science and Technology (ICIST)*. IEEE, 2024, pp. 574–582.

[21] Y. Li, G. Chen, and Z. Dong, "Multi-view graph contrastive representative learning for intrusion detection in ev charging station," *Applied Energy*, vol. 385, p. 125439, 2025.

[22] L. Xu, Z. Zhao, D. Zhao, X. Li, X. Lu, and D. Yan, "Ajsage: A intrusion detection scheme based on jump-knowledge connection to graphsage," *Computers & Security*, vol. 150, p. 104263, 2025.

[23] D. P. Kingma, M. Welling *et al.*, "An introduction to variational autoencoders," *Foundations and Trends® in Machine Learning*, vol. 12, no. 4, pp. 307–392, 2019.

[24] A. Dosovitskiy, L. Beyer, A. Kolesnikov, D. Weissenborn, X. Zhai, T. Unterthiner, M. Dehghani, M. Minderer, G. Heigold, S. Gelly *et al.*, "An image is worth 16x16 words: Transformers for image recognition at scale," *arXiv preprint arXiv:2010.11929*, 2020.

[25] Y. Meidan, M. Bohadana, Y. Mathov, Y. Mirsky, A. Shabtai, D. Breitenbacher, and Y. Elovici, "N-baiot—network-based detection of iot botnet attacks using deep autoencoders," *IEEE Pervasive Computing*, vol. 17, no. 3, pp. 12–22, 2018.

[26] F. Scarselli, M. Gori, A. C. Tsoi, M. Hagenbuchner, and G. Monfardini, "The graph neural network model," *IEEE transactions on neural networks*, vol. 20, no. 1, pp. 61–80, 2008.